\documentclass[runningheads]{llncs}
\usepackage{graphicx}
\usepackage{amsmath,amssymb,amsfonts}
\usepackage{wrapfig,lipsum,booktabs}
\usepackage{caption}
\usepackage{subcaption}
\begin{document}
\title{Efficient Out-of-Distribution Detection of Melanoma with Wavelet-based Normalizing Flows}
%
%
\author{M.M. Amaan Valiuddin\and
 Christiaan G.A. Viviers\and Ruud J.G. van Sloun\and Peter H.N. de With\and Fons van der Sommen}
%
\authorrunning{M.M.A. Valiuddin et al.}
%
\institute{Eindhoven University of Technology, Eindhoven 5612 AZ, The Netherlands
\email{m.m.a.valiuddin@tue.nl}}
\maketitle              
\begin{abstract}
Melanoma is a serious form of skin cancer with high mortality rate at later stages. Fortunately, when detected early, the prognosis of melanoma is promising and malignant melanoma incidence rates are relatively low. As a result, datasets are heavily imbalanced which complicates training current state-of-the-art supervised classification AI models. We propose to use generative models to learn the benign data distribution and detect Out-of-Distribution (OOD) malignant images through density estimation. Normalizing Flows (NFs) are ideal candidates for OOD detection due to their ability to compute exact likelihoods. Nevertheless, their inductive biases towards apparent graphical features rather than semantic context hamper accurate OOD detection. In this work, we aim at using these biases with domain-level knowledge of melanoma, to improve likelihood-based OOD detection of malignant images. Our encouraging results demonstrate potential for OOD detection of melanoma using NFs. We achieve a 9\% increase in Area Under Curve of the Receiver Operating Characteristics by using wavelet-based NFs. This model requires significantly less parameters for inference making it more applicable on edge devices. The proposed methodology can aid medical experts with diagnosis of skin-cancer patients and continuously increase survival rates. Furthermore, this research paves the way for other areas in oncology with similar data imbalance issues\footnote{Code available at: https://github.com/A-Vzer/WaveletFlowPytorch}. 
\end{abstract}
\section{Introduction}\label{sec:intro}
Melanoma, a form of skin cancer, develops in the melanocytes of the skin~\cite{melanoma1}. Symptoms can develop in the form of changing moles or growth of new pigmentation. Non-cancerous growth of the melanocytes is referred to as benign melanoma and is not harmful, while malignant melanoma is harmful. It is essential to recognize the symptoms of malignant melanoma as early as possible to classify its malignancy in order to avoid late diagnosis and ultimately an increased mortality rate~\cite{melanomaresearchalliance}. To classify melanoma malignancy, experts consider indications of the skin pigmentation such as asymmetrical shapes, irregular borders, uneven distribution of colors and large diameters (relative to benign melanoma)~\cite{mayo_clinic_2022}. These clinical properties involve characteristics related to the texture and graphical details on the skin.

Since most cases of melanoma are benign, the number of malignant melanoma images are still relatively low. This data imbalance can negatively influence the predictions of machine learning~(ML) models aiming to classify melanoma malignancy. Furthermore, most state-of-the-art supervised ML models are not calibrated, which poses the question on their validity for reliable skin-cancer detection~\cite{guo2017calibration}. Ideally, query images are assigned a calibrated confidence score, which can be interpreted as a probability of malignancy. Given these circumstances, a sensible option is to perform likelihood-based Out-of-Distribution (OOD) detection with the abundant benign data available. 

Yielding tractable distributions, Normalizing Flows (NFs) serve as an excellent method for this application. NFs are a family of completely tractable generative models that learn exact likelihood distributions. However, OOD detection with NFs is notoriously difficult. This is caused by its inherent learning mechanisms that result in inductive biases towards graphical details, such as texture or color-pixel correlations rather than semantic context in images~\cite{kirichenko2020normalizing}. As such, OOD data is often assigned similar or higher likelihoods than the training data. In this paper, we show that with domain-level understanding of melanoma, we can improve NFs for OOD detection. Since the dominant features for indicating the malignancy of melanoma are described by their size and texture, we use wavelet-based NFs. We implement Wavelet Flow~\cite{yu2020wavelet} for OOD detection of malignant melanoma and realize a 9\% performance gain in Area Under Curve (AUC) of the Receiver Operating Characteristics (ROC). The number of parameters can significantly be reduced when applying Wavelet Flow for OOD detection, enabling implementation on smaller devices.

The proposed methodology presents the potential of NFs for aiding in reliable diagnosis of melanoma. Normalizing Flows for OOD detection and its inductive biases are discussed in Section~\ref{sec:background}. Thereafter, the approach and method is discussed in Section~\ref{sec:methods}. The results are presented in Section~\ref{sec:results} and concluded in Section~\ref{sec:conclusion}.
\section{Background}\label{sec:background}
\subsection{Normalizing Flows}\label{sec:NF}
Normalizing Flows are a sequence of bijective transformations, typically starting from a complex distribution, transforming into a Normal distribution. The log-likelihood $\operatorname{log}p(\mathbf{x})$ of a sample from the Normal distribution subject to an NF transformation $f_i: \mathbb{R} \mapsto \mathbb{R}$ is computed with
\begin{equation}
    \begin{aligned}
        \log p(\mathbf{x})=\log p_\mathcal{N}\left(\mathbf{z}_{0}\right)-\sum_{i=1}^{K} \log \left(\left|\operatorname{det} \frac{d f_{i}}{d \mathbf{z}_{i-1}}\right|\right),
    \end{aligned}
    \label{eq:loglikely}
\end{equation}
where the latent sample $\mathbf{z}_i$ is from the \textit{i}-th transformation in the $K$-step NF and $p_\mathcal{N}$ the base Normal probability distribution. Due to the bijectivity of the transformations, Eq.~(\ref{eq:loglikely}) can be used to sample from $p_\mathcal{N}$ and construct a visual image with known probability. This transformation is referred to as the generative direction. An image can also be transformed in the normalizing direction (towards $p_\mathcal{N}$) to obtain a likelihood on the Normal density. Training in the normalizing direction is performed through Maximum Likelihood Estimation (MLE). Recently, many types of NFs have been proposed~\cite{rezende2015variational,durkan2019neural,kingma2016improved,huang2018neural,chen2018neural}. Better flows are generally more expressive, while having an computationally inexpensive Jacobian determinant. A widely used choice of NF are coupling flows, such as RealNVP and Glow~\cite{dinh2016density,kingma2018glow}. The latter is used as a baseline for our experiments.
\subsubsection{Out-of-Distribution detection}
The properties of NFs make them ideal candidates for OOD detection. Maximizing the likelihood of the data distribution $p(\mathbf{x})$ through a bijective transformation on $p_\mathcal{N}$ pushes away likelihoods of OOD data, when the density is normalized. Nevertheless, NFs assign similar likelihoods to train and (in-distribution) test data, indicating that flows do not overfit. This also indicates that not all OOD data receive low likelihoods. Ultimately, the assigned likelihoods are heavily influenced by the inductive biases of the model. Many NFs have inductive biases that limit their use for OOD detection applications~\cite{kirichenko2020normalizing}.
\subsubsection{Inductive biases in coupling flows}
Inductive biases of a generative model determine the training solution output and thus OOD detection performance. The input complexity plays an important role in OOD detection. Likelihood-based generative models assign lower likelihoods to more textured, rather than simpler images~\cite{serra2019input}. The widely accepted \textit{affine coupling} NF is used in this research study. Kirichenko~\textit{et al.}~\cite{kirichenko2020normalizing} show that structural parts such as edges can be recognized in the latent space. This suggests that this type of flow focuses on visual appearance such as texture and color of the images, as opposed to the semantic content. Furthermore, the authors present coupling flow mechanics that cause NFs to fail at OOD detection. This is briefly discussed in order to keep the paper self-contained, but we encourage readers to refer to the original work. Given image $\mathbf{x}$, coupling flows mask it partly ($x_m$) and update it with parameters dependent on the non-masked part $x_{res}$ as
\begin{equation}
    x_m = (x_m + t(x_{res}))\cdot e^{s(x_{res})},
    \label{eq:coupl}
\end{equation}
where $s$ and $t$ are functions that output the scale and translation parameters, respectively. The log-Jacobian determinant in Eq.~(\ref{eq:loglikely}) for coupling flows is calculated as
\begin{equation}
    \log \left(\left|\operatorname{det} \frac{d f_{i}}{d \mathbf{z}_{i-1}}\right|\right) = -\sum^D_{i=1}s_i(x_{res}),
\end{equation} 
where $i$ iterates over the image dimensionality $D$. Naturally, function $s$ is encouraged to predict high values in Eq.~(\ref{eq:coupl}) to maximize the log-likelihood in Eq.~(\ref{eq:loglikely}). To compensate for this, function $t$ must predict values that is an accurate approximation of $-x_m$. Therefore, the NF assigns high likelihood to images when the flow can accurately predict the masked part of the image. This can enable solutions that assign high likelihoods to any structured image, regardless of their semantic content. Two mechanisms are found to drive the accurate prediction of masked pixels and therefore assign higher likelihoods to OOD data. These are: learning local color-pixel correlations and information on masked pixels encoded in previous coupling layers, known as \textit{coupling layer co-adaptation}. For the latter, different masking strategies such as cycle masking can be used to deprive the model from information in previous coupling-layer iterations~\cite{kirichenko2020normalizing}. Hence, we experiment with masking strategies to counteract coupling-layer co-adaptation. As an example with the opposite effect, checkerboard masking has been proposed~\cite{dinh2016density}. Masking in this manner means that the predicted pixels are conditioned on its direct neighbouring pixels. Continuously, this encourages the NF to leverage local pixel correlations and further hinders semantically relevant OOD detection.

\subsection{Wavelet Flow}
\begin{figure}
\centering
\includegraphics[width=0.8\linewidth]{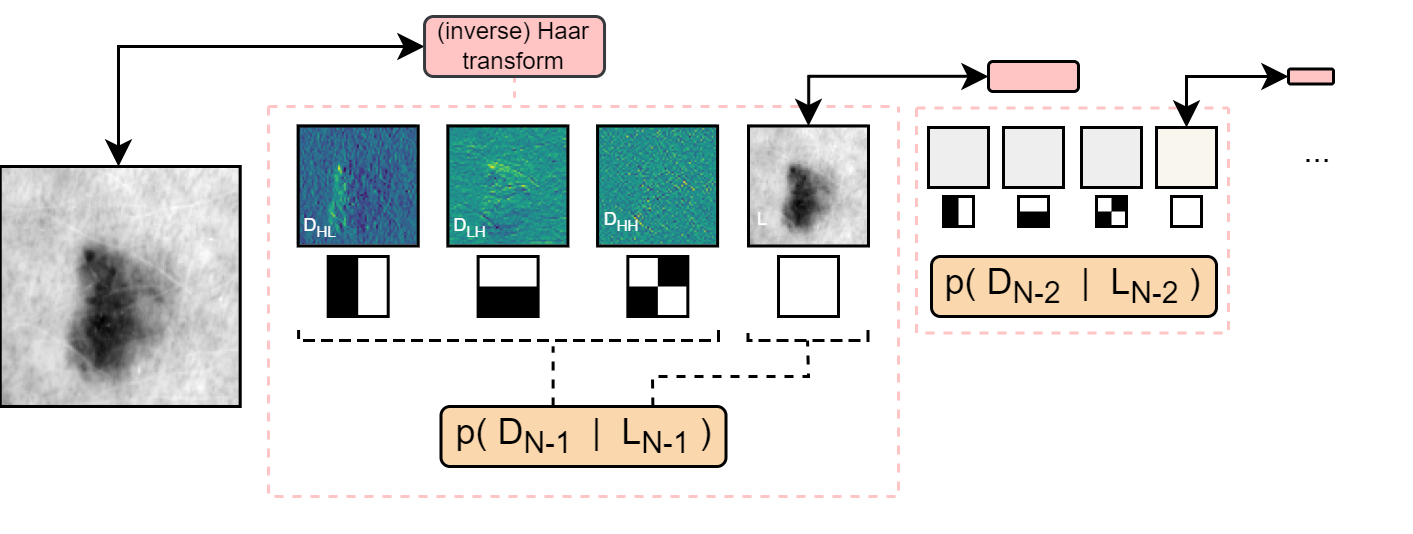}
\captionof{figure}{Wavelet Flow architecture. At each decomposition level, the likelihood of the high-frequency wavelet coefficients are learned conditioned on the low-frequency decomposition. Density $p(L_0)$ is modeled unconditionally.}
\label{fig:wfarch}
\end{figure}

Yu~\textit{et al.}~\cite{yu2020wavelet} introduced the Wavelet Flow architecture (Figure~\ref{fig:wfarch}) for efficient high-resolution image generation. Instead of learning the image pixel likelihoods, the network models the conditional distribution with a coupling NF specified by
\begin{equation}
    p(\mathbf{x})=p(L_0)\prod^{N-1}_{i=0} p(D_{i}\vert L_{i}),
\end{equation}
where $D$ and $L$ are the detail and low-frequency components of the Haar decomposition, respectively, and $N$ represents the number of decompositions. During inference, an independent sample from $p(L_0)$ is up-scaled with the inverse Haar transform, using the predicted wavelet coefficients. To the best of our knowledge, this architecture has not been tested for OOD detection. Modeling the wavelet coefficients further guides the model to consider the graphical details of the image. As discussed in Section~\ref{sec:intro}, melanoma can be distinguished by the texture of the skin. As a result, this inductive bias can improve OOD detection of melanoma. Furthermore, the high-frequency (detail) coefficients of the image enable easier distinguishment between highly textured malignant and less structured benign melanoma. This can facilitate better OOD detection, as NFs tend to assign higher likelihoods to smoother images.
\section{Methods}\label{sec:methods}
\begin{figure}
\centering
\includegraphics[width=0.85\linewidth]{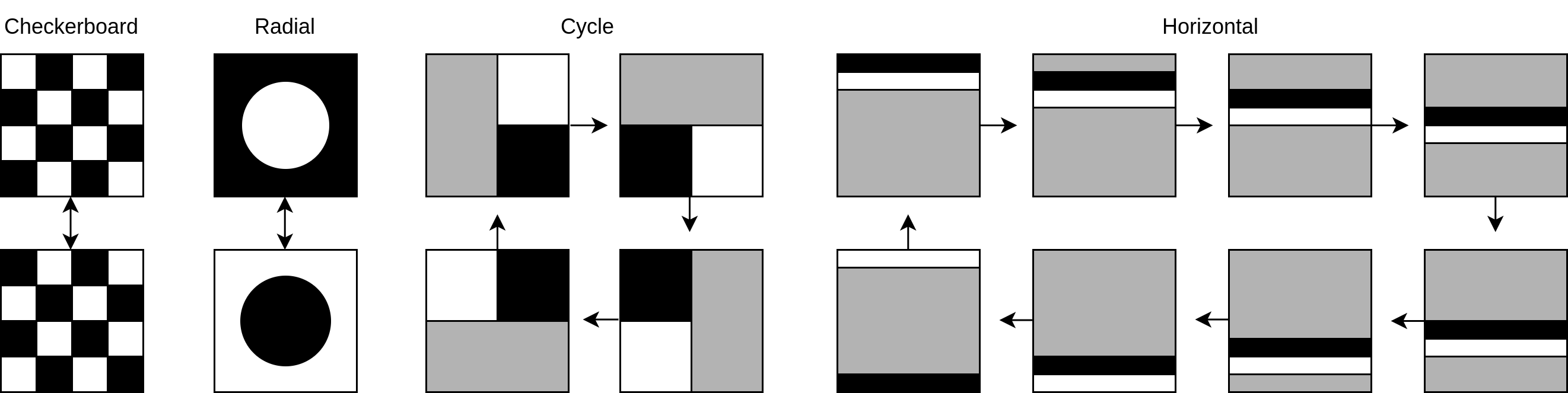}
\captionof{figure}{Various masking strategies. The masks vary at each coupling flow step. The white area indicates the input of the $s,t$-network, which predicts parameters for the masked area in black. Grey areas are disregarded in the coupling process.}
\label{fig:masking}
\end{figure}

As discussed in Section~\ref{sec:NF}, the inductive biases of coupling NFs restrict their OOD detection capabilities. Given this information, we improve this by changing the data and model architecture. We test our approach on the ISIC dataset~\cite{codella2018skin}. In this case, it can be beneficial that generative models assign higher likelihoods to less complex images, because benign melanoma are less textured and smaller in radius~\cite{mayo_clinic_2022}. Initially, we downscale the RGB images to 128$\times$128 pixels and train on the GLOW architecture naively, in a multi-scale setting, with default parameters $K=32$ and $L=3$. The AUC of the ROCs are used to evaluate the model performances. The color channels are heavily correlated and influence the likelihoods adversely, as discussed in Section~\ref{sec:NF}. Therefore, we use grayscale images to hinder exploitation of local color-pixel correlations as well as to reduce training complexity. Thereafter, Wavelet Flow is employed. This shifts the optimization from the image pixels to their wavelet coefficients. This will further bias the model towards the graphical appearance of the images, since the tumor malignancy will be even more distinguishable by texture. Additionally, we experiment with different masking strategies (see Figure~\ref{fig:masking}). With Wavelet Flow, we obtain a likelihood, and thus an AUC score per decomposition scale. The individual likelihoods are averaged over all scales that contain sufficient information about the original content of the image. In this case, these are wavelet coefficients from 4$\times$4 pixel dimensions up until the highest decomposition level. It might be beneficial to select only particular scales with good AUC values. However, this would constitute supervision, i.e. access to the malignant class, which is beyond the scope of this paper. 

\newpage
\section{Results and discussion}\label{sec:results}
\begin{figure}
\centering
\vspace{2em}\includegraphics[width=0.9\linewidth]{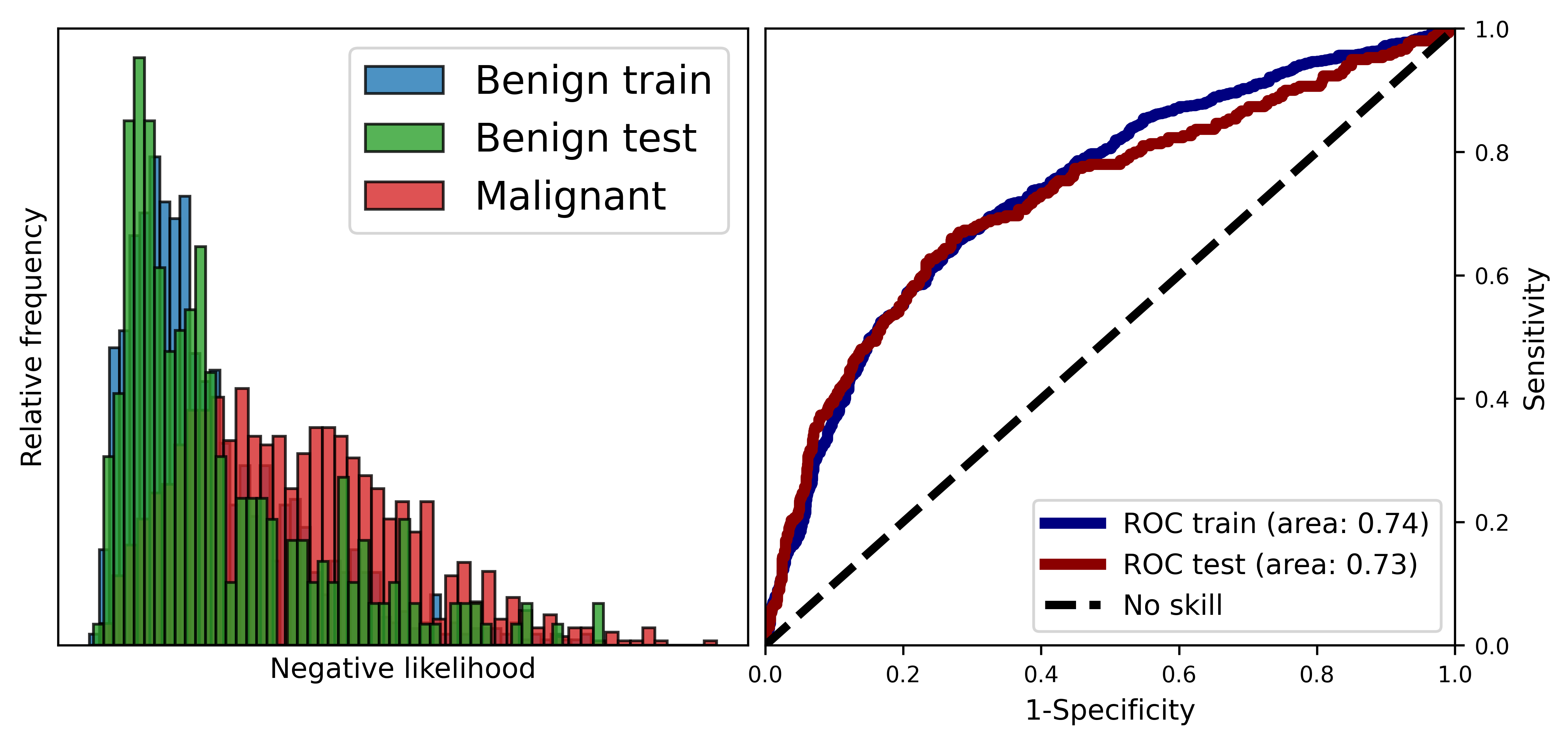}
\captionof{figure}{Likelihood distribution and ROC curve of the trained GLOW architecture}
\label{fig:glow}
\end{figure}

Table~\ref{tab:results} presents the ROC curves for the various tested models. Likelihood distributions of the GLOW architecture trained on color images are shown in Figure~\ref{fig:glow}. Firstly, it can be observed that the train and test sets coincide well, indicating the absence of overfitting. When comparing the benign test to the malignant likelihoods, we obtain an AUC of 0.73. This solution is sub-optimal because many benign images were assigned low-likelihood scores. In the same likelihood range, most of the malignant images are present as well. This is because the model learns color-pixel correlations which can be used to leverage accurate predictions of the masked latent variables in the coupling layers. As a result, this leads to higher likelihoods assigned to OOD data.

\begin{figure}
\centering
\includegraphics[width=0.9\linewidth]{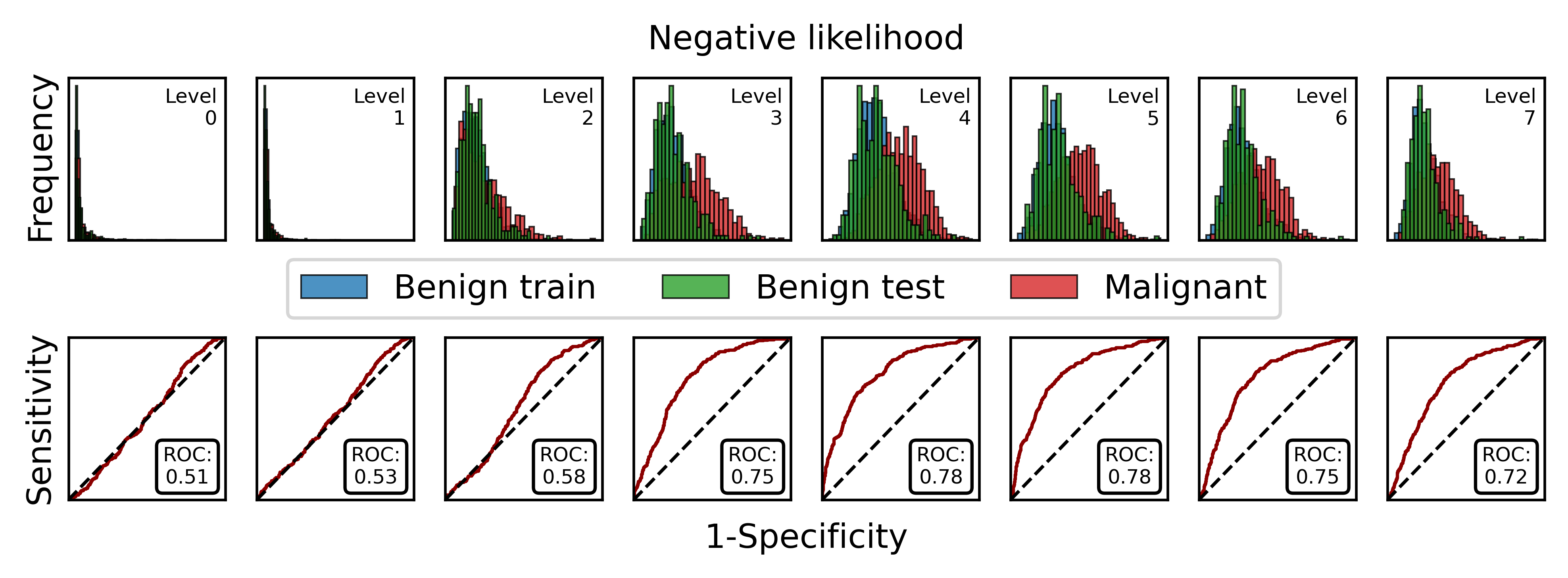}
\captionof{figure}{Likelihood distributions per Haar wavelet decomposition level}
\label{fig:wf1}
\end{figure}

\begin{figure}
\centering
\vspace{2em}\includegraphics[width=0.9\linewidth]{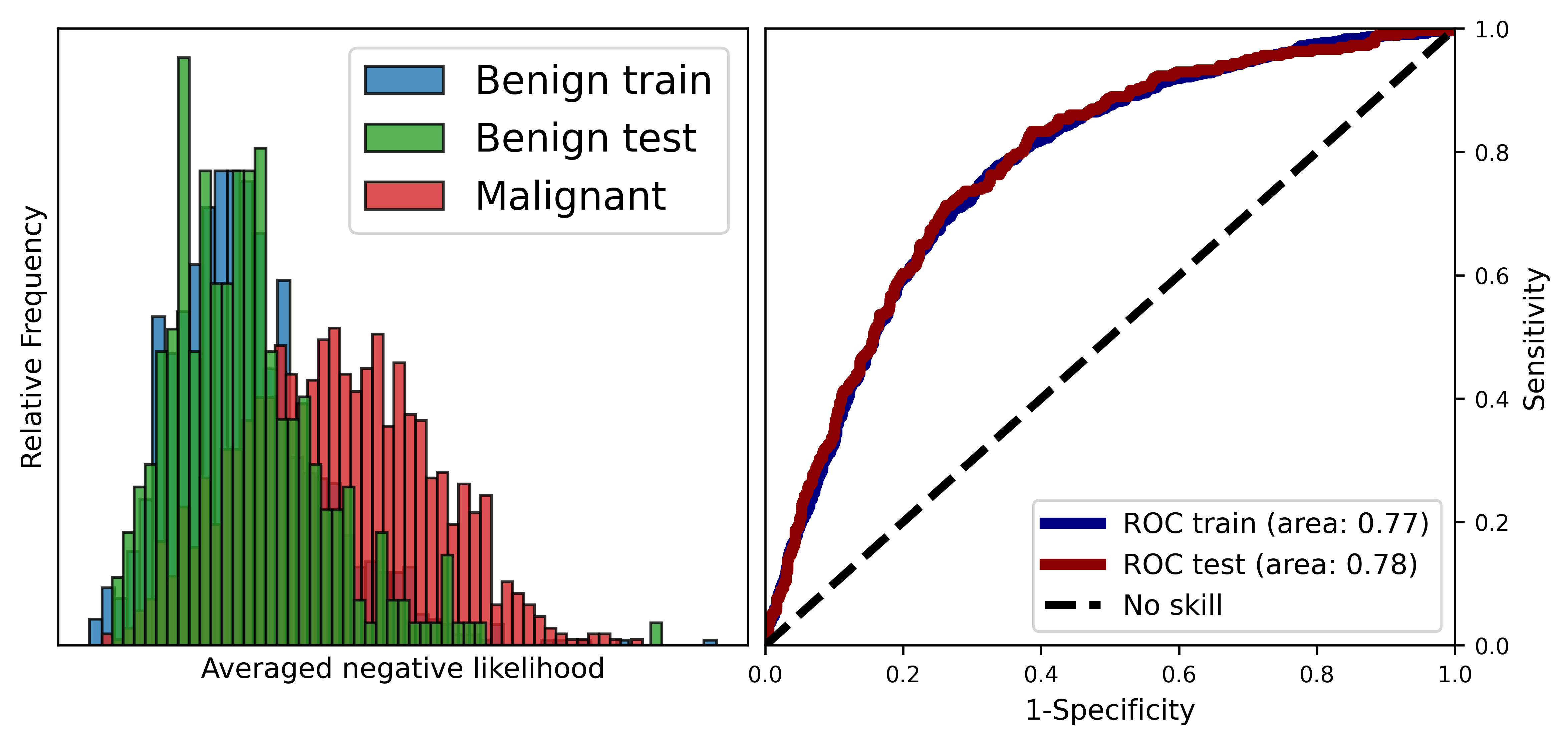}
\captionof{figure}{The likelihood distribution and ROC curve of the trained Wavelet Flow architecture, averaged over the decomposition scales}
\label{fig:wf2}
\end{figure}

When training on the wavelet coefficients with Wavelet Flow, there is substantial improvement on several decomposition scales (see Figure~\ref{fig:wf1}). At all of the decomposition scales, besides the level seven (corresponding to the highest image resolution), we observe an improvement in test evaluation. We find the best AUC values from the 3rd up until the 6th decomposition scales. At these levels, the wavelet coefficients represent the most relevant frequency components of benign and malignant melanoma. As expected, the lowest decomposition scales contain almost no relevant information on the malignancy of melanoma and have very low AUC values. In Figure~\ref{fig:wf2}, we average the likelihoods over the relevant decomposition scales. In a separate evaluation, we performed OOD detection using only the magnitude of the wavelet coefficients in which we observed acceptable AUC values on individual scales. However, in contrast with Wavelet Flow, averaging over the decomposition scales worked adversely. Furthermore, the different masking strategies did not improve performance.
\begin{table}[b]
\centering
\setlength{\tabcolsep}{0.7em} 
\begin{tabular}{lccccccc} 
\toprule
\textbf{Architecture} & \textbf{K} & \textbf{L} &\textbf{channels} & \textbf{masking} & \textbf{ROC} & \textbf{\# parameters*} \\
\hline
GLOW &32&3& RGB & Affine & 0.73 & 159M\\
GLOW &32&3& Gray & Affine & 0.74 & 9.51M \\
GLOW &32&1& RGB & Affine & 0.72 & 3.47M\\
GLOW &32&1& Gray & Affine & 0.75 & 2.57M \\
Wavelet Flow &32&1& Gray & All & 0.78 & 2.50M\\
Wavelet Flow &16&1& Gray & All & \textbf{0.78} & \textbf{1.25M}\\
\bottomrule
\end{tabular}
    \caption{Test set results of the models trained on the ISIC dataset. For Wavelet Flow the number of parameters are that of the highest decomposition level as each level can be trained independently.}\label{tab:results}
\end{table}

\begin{figure}[t]
\centering
\includegraphics[width=0.9\linewidth]{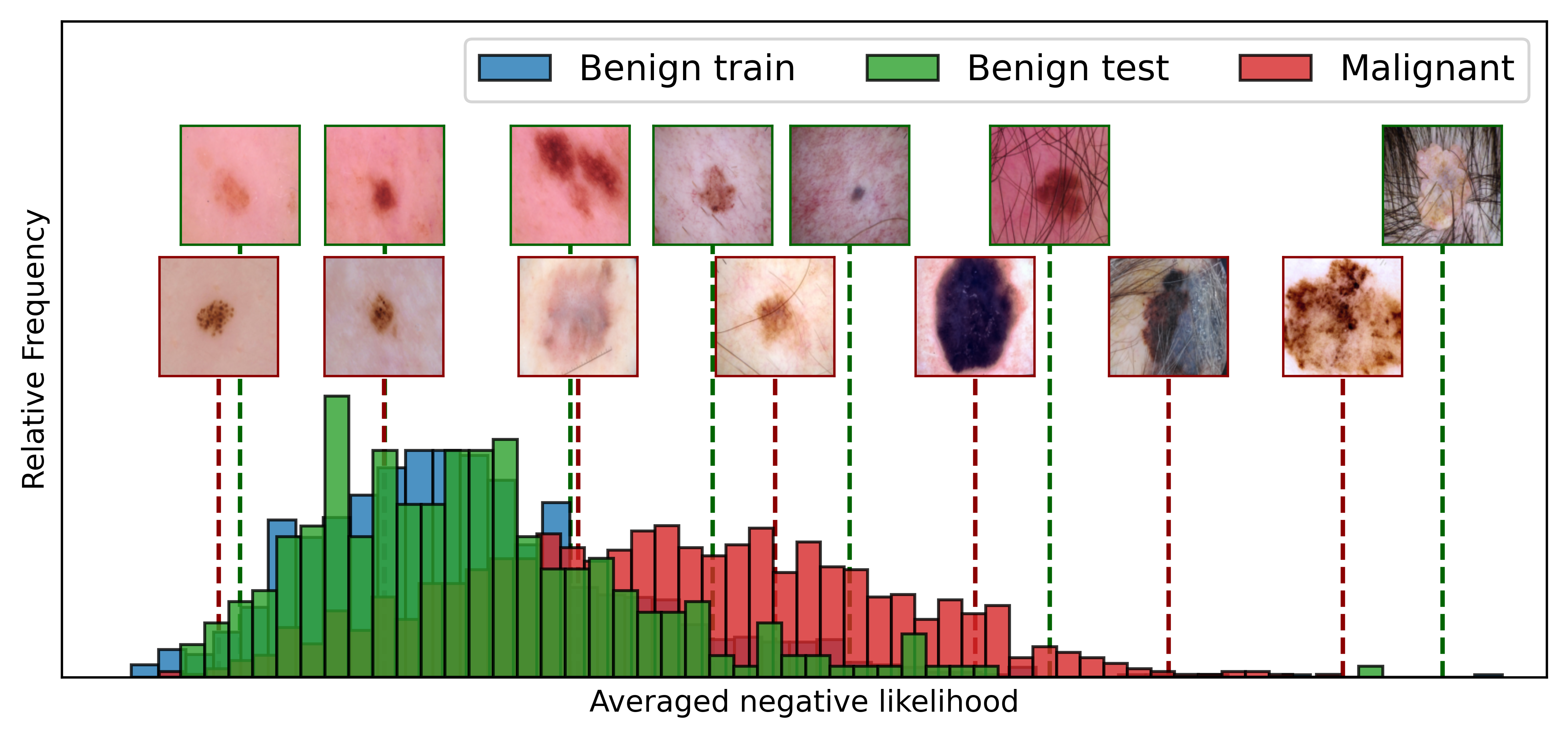}
\captionof{figure}{Images of benign and malignant melanoma at various likelihoods. Note that lower likelihoods are either malignant or highly textured benign melanoma.}
\label{fig:wf3}
\end{figure}

Finally, some images are depicted of benign and malignant images of melanoma around various likelihoods (Figure~\ref{fig:wf3}). We notice that at higher likelihoods, the malignant samples are more similar to that of the benign images. This indicates that the model is sufficiently learning relevant features, but is unable to classify early-phase malignant melanoma. As the likelihood values decrease, we observe more textured images. Specifically, larger pigmentation is visible together with more hairs. The hairs increase the activations in the wavelet domain, augmenting image complexity, resulting in lower likelihoods. The likelihood calculations can be corrected with a complexity term that considers hairiness, similar to Serr\`{a}~\textit{et al.}~\cite{serra2019input}. This will shift hairy benign images to higher likelihoods. We leave the implementation of this correction term for further work. For malignant melanoma, it can be seen that the consideration of texture goes beyond hairiness and size of pigmentation. Large skin pigmentations with minimal texture are more likely to be benign. This indicates yet again that the inductive biases of Wavelet Flow cause the model to sufficiently extract relevant information from the images.
\section{Conclusion}\label{sec:conclusion}
Late diagnosis of melanoma poses high risks for patients with skin cancer. Early detection of malignant melanoma with machine learning is highly valuable, but is difficult due to data imbalance caused by its relatively low occurrence. We learn the benign image data distribution with Normalizing Flows to perform Out-of-Distribution (OOD) detection. We show that with knowledge on melanoma and the inductive biases of Normalizing Flows, we can improve likelihood-based OOD detection with wavelet-based Normalizing Flows. Furthermore, we demonstrate that memory requirements for OOD detection can significantly be reduced with Wavelet Flow, enabling the deployment on edge devices. We recommend including a term in the likelihood calculations that correct for presence of hairs in future work. The proposed methodology focuses solely on melanoma, however, we suggest that further research can facilitate exact likelihood-based OOD detection for other areas of oncology with large data imbalances to improve detection accuracy.
\newpage
\bibliographystyle{splncs04}
\bibliography{refs}
\end{document}